\documentclass{article}

\usepackage{arxiv}

\usepackage[utf8]{inputenc} % allow utf-8 input
\usepackage[T1]{fontenc}    % use 8-bit T1 fonts
\usepackage{hyperref}       % hyperlinks
\usepackage{url}            % simple URL typesetting
\usepackage{booktabs}       % professional-quality tables
\usepackage{amsfonts}       % blackboard math symbols
\usepackage{nicefrac}       % compact symbols for 1/2, etc.
\usepackage{microtype}      % microtypography
\usepackage{lipsum}		% Can be removed after putting your text content
\usepackage{graphicx}
\usepackage[numbers]{natbib}
\usepackage{doi}

\usepackage{amsmath}
\usepackage{amssymb}
\usepackage{bbm}  % for \mathbbm

\usepackage{algorithm}
\usepackage{algpseudocode}

\title{Generative Learning of Separatrices}

\author{ \href{https://orcid.org/0000-0000-0000-0000}{\includegraphics[scale=0.06]{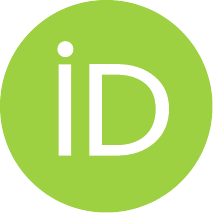}\hspace{1mm}Ellis R. Crabtree}\\%\thanks{Use footnote for providing further
	Dept. of Predictive Analytics\\
	Johns Hopkins All Children's Hospital\\
	St. Petersburg, FL 33701 \\
	\texttt{ellis@jhmi.edu} \\
	\And
	\href{https://orcid.org/0000-0000-0000-0000}{\includegraphics[scale=0.06]{orcid.pdf}\hspace{1mm}Dimitris G. Giovanis} \\
	Dept. of Civil and Systems Engineering\\
	Johns Hopkins University\\
	Baltimore, MD 21218\\
	\texttt{dgiovan1@jhu.edu} \\
    \And
	\href{https://orcid.org/0000-0000-0000-0000}{\includegraphics[scale=0.06]{orcid.pdf}\hspace{1mm}Anastasia Georgiou} \\
	Dept. of Chemical and Biomolecular Engineering\\
	Johns Hopkins University\\
	Baltimore, MD 21218\\
	\texttt{ageorgi3@jhu.edu} \\
    \And
	\href{https://orcid.org/0000-0002-6427-2385}{\includegraphics[scale=0.06]{orcid.pdf}\hspace{1mm}George Datseris} \\
	Dept. of Mathematics and Statistics\\
	University of Exeter\\
	Exeter, EX4 4QJ\\
	\texttt{g.datseris@exeter.ac.uk} \\
    \And
	\href{https://orcid.org/0000-0000-0000-0000}{\includegraphics[scale=0.06]{orcid.pdf}\hspace{1mm}Ioannis G. Kevrekidis} \\
	Dept. of Chemical and Biomolecular Engineering\\
	Johns Hopkins University\\
	Baltimore, MD 21218\\
	\texttt{yannisk@jhu.edu} \\
}

\renewcommand{\shorttitle}{Generative Learning of Separatrices}

\hypersetup{
pdftitle={Generative Learning of Separatrices},
pdfauthor={Ellis R.~Crabtree, Dimitris G.~Giovanis, Anastasia ~Georgiou, George Datseris, Ioannis G. Kevrekidis},
pdfkeywords={Generative Models, Uncertainty Quantification, Dynamical Systems},
}

\begin{document}
\maketitle

\begin{abstract}
The identification and reconstruction of the boundaries separating basins of attraction in multistable, multidimensional dynamical systems presents a fundamental challenge in computational dynamics. These (often geometrically complex) structures govern transition pathways and other important large timescale behavior, yet they remain typically under-sampled since their neighborhood does not get routinely visited during direct simulations. Traditional computational approaches targeted for their approximate construction (including manifold continuation, edge tracking, and bisection algorithms, among others) face computational limitations in high-dimensional systems and require \emph{a priori} knowledge of the dynamical system and its equations. Simplistic sampling methods such as random or uniform sampling of the phase space typically fail to quantitatively approximate separatrices and their structure altogether, suffering increasingly from the curse of dimensionality as the system dimension grows.

We introduce and implement a framework that combines {\em supervised classification} with {\em generative modeling} to address this challenge. Our approach first trains neural network classifiers on uniformly or randomly sampled initial conditions, importantly labeled by their corresponding basins of attraction in the system of interest. Using uncertainty metrics of the trained classifier to quantify decision boundaries, the method then identifies these high uncertainty regions and boundaries of the classifier as preliminary {\em approximate separatrices}. Subsequently, score-based generative models are trained specifically on samples from high-uncertainty regions, ultimately generating densities of samples consistent with the empirical density of samples on the manifold (or regions close to this manifold) that constitutes the separatrix between basins in the sampled region. This approach leverages the complementary strengths of (a) discriminative models for global phase space partitioning and (b) generative models for detailed geometric sampling, resulting in a systematic, iterative, data-driven framework that produces empirically consistent reconstructions of (approximate) separatrix manifolds.
\end{abstract}

% keywords can be removed

\section{Introduction}
\label{sec:intro}
A valuable step in the modeling of multistable systems is the characterization of the boundaries of their basins of attraction, commonly known as \emph{separatrices}. These separatrices often possess intricate and complex geometric structures that may even be fractal in nature. Typically, these separatrices correspond to stable manifolds of saddle-type invariant sets, unstable (source-type) invariant sets such as limit cycles, or other complex manifolds.
In multistable dynamical systems~\cite{Feudel2018MultistabilityTippingMathematics, Pisarchik2022} trajectories initialized from arbitrary, generic points in phase space tend to converge to one of several attracting sets—fixed points, limit cycles, strange attractors, or other attracting invariant manifolds. The basin of attraction for each attractor comprises the set of initial conditions whose forward-time trajectories asymptotically approach that attractor~\cite{Datseris2022NonlinearDynamicsJulia}. Long-time numerical simulations of said systems often experience difficulty in sampling the phase space of these systems in sufficient detail because their trajectories become effectively trapped in neighborhoods of attractors, yielding statistically stationary behavior that provides information only about the local dynamics within a single basin.

Characterizing separatrices presents significant computational and theoretical difficulties that have motivated substantial research in computational dynamics and adjacent fields. 
In principle, if one has a dense coverage of estimated basins via emergent techniques such as the \emph{recurrences-based}~\cite{Datseris2022BasinsAttraction} or \emph{featurize-and-group}~\cite{Datseris2023FrameworkGlobalStability} methods, identifying the separatrix may be tractable.
This however becomes quickly infeasible as system dimension increases, as the cost of a dense covering of the basins scales exponentially.
Sampling points near the separatrices in order to enhance the time computational orbits spent near them is equally challenging, as separatrices typically correspond to state space regions of low probability density and often have formally zero volume.
This leads to the problem that the regions that govern transitions between basins and determine the system's response to perturbations are those least visited by typical trajectories.

Recent advances in applied mathematics have yielded several complementary approaches for identifying and approximating separatrices in high-dimensional dynamical systems. These include techniques such as manifold continuation \cite{krauskopf2005survey, aronson1982, kevrekidis1987bifurcations, johnson1997two}, 
edge tracking and bisection algorithms \cite{skufca2006edge}, or Lyapunov exponent Monte Carlo sampling \cite{Armiyoon2014}. The application of these and more rudimentary techniques spans diverse fields including chemical kinetics and reactor design (identifying steady states and bifurcations in chemical reactions) \cite{UPPAL1974967, ELNASHAIE19931}, ecology (regime shifts in population dynamics and ecosystems) \cite{scheffer2001catastrophic}, climate science \cite{feudel1997multistability}, and even comprehensive, extremely high dimensional climate models \cite{Boerner2025}. Reconstructing separatrices provides not only a characterization of critical transition thresholds but also generates representative samples from regions of phase space that are otherwise markedly undersampled by direct numerical simulation, while the prior mentioned methods such as continuation require explicit knowledge of equations.

In parallel, the machine learning community has developed powerful generative modeling frameworks capable of learning to sample from complex probability distributions from finite sample sets and subsequently generating new samples that reflect the statistical properties of the training data, with early work involving models such as variational autoencoders and generative adversarial networks \cite{VAE, goodfellow2014}. Score-based generative models (SGMs) represent a particularly promising class of deep generative models \cite{songdiffmodels, diffmodels2}. These models learn the score function $\nabla_x \log p(x)$, where $p(x)$ is the target data distribution, and use this learned score to generate samples via Langevin dynamics or by solving a reverse-time stochastic differential equation. Additionally, these architectures have been applied to climate modeling problems specifically targeting the generation of samples near tipping points and separatrix regions \cite{sleeman2023generative}, demonstrating that deep generative models can be specialized to focus on dynamically critical regions of phase space. 
Crucially, recent work has identified connections between generative models such as SGMs and classical techniques in statistical physics and enhanced sampling \cite{GANs_closures, crabtree2024micro}. Furthermore, SGMs and other generative models have been shown to generate density-consistent data on manifolds \cite{giovanis2025generative, crabtree2025generative}, and have demonstrated state-of-the-art performance in high-dimensional generative tasks, and they exhibit a remarkable ability to recover low-dimensional manifold structure embedded in high-dimensional ambient spaces \cite{diffmanifolds}. This has enabled hybrid approaches that combine the representational power of deep neural networks with physics-informed constraints and sampling strategies.

As a continuation/extension of these efforts, this work presents a framework that combines (a) supervised classification of basins of attraction with (b) sampling of densities on a manifold performed by a generative model, to achieve comprehensive characterization and reconstruction of separatrices in multistable dynamical systems. The methodology proceeds as follows:

\begin{enumerate}
    \item \textbf{Basin classification:} Train neural network classifiers to predict basin membership from provided phase space coordinates with known (labeled) basins.
    
    \item \textbf{Decision boundary identification:} Extract the decision boundaries of the trained classifier with the help of uncertainty metrics; these boundaries will approximate the true separatrices under appropriate conditions.
    
    \item \textbf{Generative modeling on boundaries:} Train score-based or other generative models specifically on data of high uncertainty, which lie near the identified decision boundaries, enabling targeted sampling of separatrix regions.
    
    \item \textbf{Refinement and visualization:} Use generated samples to refine separatrix approximations iteratively and produce high-fidelity visual representations of basin boundary geometry.
\end{enumerate}

This integrated approach leverages the complementary strengths of discriminative and generative models: classification provides global phase space partitioning and basin boundary localization, while generative models guide (and thus enhance) detailed sampling and reconstruction of the geometric structures comprising those boundaries. The framework is particularly well-suited to high-dimensional systems where traditional continuation or bisection methods become computationally prohibitive, and provides a data-driven (and density-consistent with the empirical data) pathway to extracting geometric structures of separatrices from simulation or experimental time series data.

\section{Proposed Mathematical Framework}
\label{sec:algorithm}

Consider an autonomous dynamical system in $\mathbb{R}^n$ governed by a flow $\Phi^t(\mathbf{x}) = \mathbf{x}(t)$. Assume that this system contains $k$ distinct attractors $A_1, A_2, \ldots, A_k$ with corresponding basins of attraction $B_1, B_2, \ldots, B_k$. The basin boundaries $\partial B_i$ are $(n-1)$-dimensional sets (generically) that coincide with stable manifolds $W^s(S)$ of saddle-type invariant sets $S$. These saddle points, unstable periodic orbits, or chaotic saddles play a critical role in organizing the phase space topology and mediating transitions between basins under perturbation.

Our proposed approach uses a classification neural network to classify (assign the correct) basins of attraction $B_i$ for given initial conditions. Beyond the assignment of the right basin for a given point, the classifier also reports {\em an uncertainty metric} (i.e. how sure the network is that the predicted basin is correct). The uncertainty metric used in this work is based on Monte Carlo dropout \cite{gal2016dropout}, which is represented by an entropy. This entropy, $H$, is a measure of how much the network's predictions vary when it is run multiple times with different random subsets of neurons temporarily switched off (dropout); if the predictions consistently agree, H is low and the classifier is confident, but if they disagree, H is high, signaling that the point likely lies in an uncertain region—such as near a boundary between basins.

A generative model is then given initial conditions as training data with the uncertainty metric as a label; this then allows for the generation of new points at high uncertainty phase space areas, which will naturally lie at or near boundaries between the basins of the system. These high uncertainty points function as proxy sampling points for local and global separatrices and/or their surrounding regions of phase space, depending on sampling density and model parameters. Algorithms 1 and 2 outline the framework.

\begin{algorithm}
\caption{Basin Classifier Network Training and Inference}
\textbf{Input:} Training data $\mathcal{D}_1 = \{(x(t_0)_i, B_i)\}_{i=1}^N$ (initial conditions from a dynamical system, $x(t_0)$, and discrete labels given by the initial condition's basin of attraction, $B_i$ that is reached from each initial condition after some time $t$), neural network $f(\mathcal{D}_1); \theta)$ with dropout layers, loss function $\mathcal{L}$, number of epochs $E$, batch size $B$, number of MC samples $M$

\label{alg:Training_Classifier}
\begin{algorithmic}[1]
\Statex \textbf{--- Training Phase ---}
\For{epoch $= 1$ to $E$}
    \State Shuffle training data
    \For{each batch $\{(x_b, y_b)\}_{b=1}^B$}
        \State Enable dropout
        \State Compute predictions: $\hat{y}_b = f(x_b; \theta)$
        \State Compute loss: $\mathcal{L}_{\text{batch}} = \mathcal{L}(\hat{y}_b, y_b)$
        \State Update parameters: $\theta \gets \theta - \eta \nabla_\theta \mathcal{L}_{\text{batch}}$
    \EndFor
\EndFor

\Statex \textbf{--- Inference Phase (Monte Carlo Dropout) ---}
\Require New input sample $x(t_0)$
\State Initialize average probabilities, $\bar{p}$ as a zero vector
\For{$m = 1$ to $M$}
    \State Enable dropout at inference
    \State Run forward pass of network: \texttt{$\hat{y}_m$} $\gets$ $f(x(t_0)_m)$
    \State Apply softmax to get probabilities: \texttt{$p_m$} $\gets$ softmax(\texttt{$\hat{y}_m$})
    \State Get average probabilities: $\bar{p} \gets \bar{p} + \frac{1}{M} p_m$
\EndFor
\State Compute uncertainty (entropy): $H(\bar{p}) = -\sum_m \bar{p}_k \log \bar{p}_k$
\State \Return $\hat{y}, \bar{p}, H(\bar{p})$
\end{algorithmic}
\end{algorithm}

\begin{algorithm}
\caption{Training a score-based generative model (SGM) to approximate the separatrices}
{\textbf{Input:} Training data $\mathcal{D}_2 = \{(x(t_0)_j)\}_{j=1}^N$ (initial condition samples from a dynamical system, $x(t_0)$, with corresponding distribution $P(\mathcal{D}_2)$, where $\mathcal{D}_2$ is chosen by selecting an entropy threshold from the Monte Carlo dropout of the classifier network, $H(\bar{p}$). For practical notation purposes we will call the uncertainty threshold $z$.}
\label{alg:Training_SGM_alg}
\begin{algorithmic}[1]
\State Retain high entropy $\mathcal{D}_2$ data from Algorithm \ref{alg:Training_Classifier}. In this work, all initial conditions with $z$ above the 90th percentile were retained.
\State Train a SGM on $\mathcal{D}_2$ to model the distributions of high entropy samples
\State Use the trained SGM to sample from $\hat{P}(\mathcal{D}_2)$ — approximated measures in ambient space at the prescribed high entropy \\

\Return $\hat{P}(\mathcal{D}_2)$ for high $z$
\end{algorithmic}
\end{algorithm}

The identification of separatrices with high-uncertainty regions from the trained classifier represents an approximation whose validity depends critically on the relationship between the classifier's learned decision boundary and the true dynamical basin boundary. Under ideal conditions—sufficient training data uniformly distributed across phase space, adequate network capacity, and successful optimization—the classifier learns to assign high probability to the correct basin membership; classification uncertainty (quantified via entropy of the MC dropout predictive distribution) should peak near the true separatrix $\partial B_i$, where infinitesimal changes in initial conditions lead to different asymptotic outcomes. However, several sources of systematic bias affect this approximation. First, the classifier decision boundary depends on the training data distribution. Regions of phase space with sparse coverage of sampled initial conditions will exhibit elevated uncertainty regardless of proximity to the true separatrix, since the network has insufficient information to make confident predictions. This creates a fundamental confound: high uncertainty may indicate either proximity to a basin boundary or absence of training data. 
If the model is not trained on sufficiently numerous examples near the true decision boundary, and that boundary is not geometrically simple, the model may confidently misclassify instances by erroneously extrapolating learned patterns beyond their valid range.

Furthermore, the geometry of the classifier decision boundary is constrained by the neural network's inductive bias. Standard feedforward networks with smooth activation functions (e.g., ReLU, tanh) produce piecewise smooth decision boundaries whose complexity scales with network depth and width. For separatrices with fractal structure or those involving fine-scale geometric detail, finite networks necessarily produce smoothed approximations. The decision boundary may capture large-scale topology while missing fine structure below the effective resolution determined by network architecture and training procedure. Additionally, training data imbalance between basins systematically biases decision boundaries. If basin $B_1$ is overrepresented relative to basin $B_2$ in the training set, standard cross-entropy loss could incentivize the classifier to expand the predicted region for $B_1$ at the expense of $B_2$, shifting the decision boundary away from its dynamically correct location. This is particularly problematic when the dynamics itself creates sampling bias.

The approximation is most reliable when: (1) training trajectories provide relatively uniform coverage of phase space, perhaps through strategic initial condition selection, (2) sufficient trajectory data exists near the separatrix, which can be achieved iteratively by generating initial conditions (and subsequent trajectories) from high-uncertainty regions, (3) the network architecture is sufficiently expressive to capture the separatrix geometry at the scale of interest, and (4) class balance is maintained or explicitly corrected through loss function weighting. These considerations motivate the iterative refinement approach and the integration of generative models: by producing samples near/along provisional boundary estimates, the generative model enables targeted generation of new trajectories that reduce sampling bias, while validation against dynamical properties (e.g., checking that generated points lie near saddle points/manifolds or exhibit long transient times) provides quality control on the reconstruction. To this point, the entropy threshold selected can offer some fine control on how "wide or narrow" the end result approximation of the separatrix becomes. In this work, a threshold of ninety percent was decided to be quantitative and qualitatively sufficient for the examples that follow.
\section{Numerical Examples}
\label{sec:examples}

We demonstrate the proposed framework on three dynamical systems that exhibit multistability with progressively complex basin boundary structures. The Newton method fractal boundary provides a two-dimensional benchmark with intricate, self-similar basin boundaries whose fractal geometry is well-characterized, allowing us to assess the method's ability to reconstruct fine-scale geometric detail. The continuous stirred tank reactor (CSTR) model represents a canonical problem in chemical process engineering where separatrices delineate operationally critical boundaries between stable regimes, and where traditional continuation methods have been extensively applied. Finally, the three-dimensional Lorenz system exemplifies chaotic dynamics where basin boundaries, formed by two-dimensional stable manifolds of saddle equilibria, separate trajectories leading to qualitatively different asymptotic behaviors, possibly on a strange attractor. Together, these examples demonstrate the framework's capability to handle systems with varying dimensionality, geometric complexity, and practical significance, while demonstrating advantages over traditional methods by sidestepping computationally intensive integration/continuation and avoiding any required \emph{a priori} knowledge of the dynamical system.

\subsection{The Newton Method Fractal Boundary}

\begin{figure}[H]
    \centering
    \includegraphics[width=0.7\linewidth]{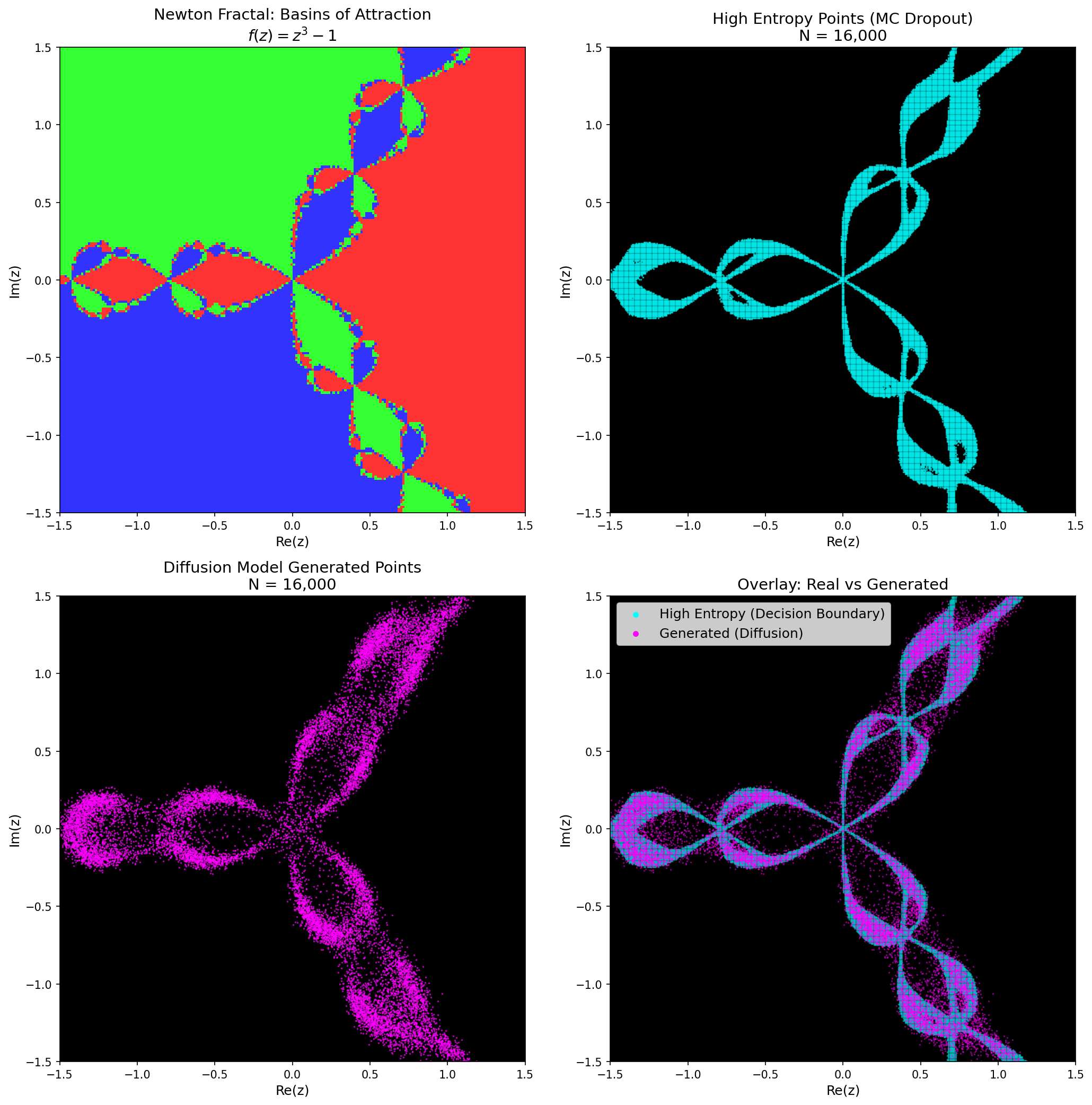}
    \caption{Newton fractal basin classification for the complex polynomial $f(z) = z^3 - 1$, illustrating the uncertainty quantification and generative sampling framework across four subfigures. \textbf{Top-left:} uniformly sampled initial conditions in the complex plane, labeled and colored by their basin of attraction under Newton's method iteration $z_{n+1} = z_n - f(z_n)/f'(z_n)$. \textbf{Top-right:} high-entropy points identified by the UQ metric derived from the classifier neural network, highlighting regions of fractal boundary ambiguity where basin membership is uncertain. \textbf{Bottom-left:} generated samples produced by a score-based generative model (SGM) trained exclusively on the high-entropy initial conditions, learning to reproduce the complex boundary structure. \textbf{Bottom-right:} overlay comparison in the complex plane contrasting the true high-entropy data against the SGM-generated samples, demonstrating the model's ability to capture the fractal geometry near basin boundaries.}
    \label{fig:fractal-main}
\end{figure}

The Newton fractal is generated by applying Newton's method for root finding to the polynomial $f(z) = z^3 - 1$ in the complex plane. The iterative map is given by

\begin{equation}
    z_{n+1} = z_n - \frac{f(z_n)}{f'(z_n)} = z_n - \frac{z_n^3 - 1}{3z_n^2} = \frac{2z_n^3 + 1}{3z_n^2},
    \label{eq:Newton}
\end{equation}

where $z \in \mathbb{C}$. Each initial condition $z_0 \in \mathbb{C}$ converges to one of three roots: $r_1 = 1$, $r_2 = e^{i2\pi/3}$, and $r_3 = e^{i4\pi/3}$. The basins of attraction for these roots are separated by fractal boundaries. This system provides an interesting test case for validating separatrix reconstruction accuracy due to its well-studied geometric properties and the self-similar structure of its basin boundaries across multiple spatial scales due to its fractal nature. Figure \ref{fig:fractal-main} demonstrates the algorithmic framework presented in section \ref{sec:algorithm} in three steps: basin classification, using UQ metrics to identify the classifier decision boundary given by the high entropy initial conditions, and generative sampling of the separatrix by an SGM trained on the high entropy initial conditions. Forty thousand uniformly sampled initial conditions (a 200x200 mesh grid) on the complex plane were integrated 50 steps each using Newton's method using equation \ref{eq:Newton}. This grid of initial conditions labeled by the endpoints of their trajectories were used to train the neural network classifier. Once trained, the classifier evaluated a new finer mesh grid of 160,000 (400x400) samples. The top ten percent of the evaluated samples with respect to their entropy predicted by the classifier (16,000 samples) were retained for SGM training. Once trained on the high entropy points, the SGM then generated 16,000 new samples for comparison with the high entropy training dataset that effectively approximates the decision boundary (and thus the separatrix) of the system.

\begin{figure}[H]
    \centering
    \includegraphics[width=0.9\linewidth]{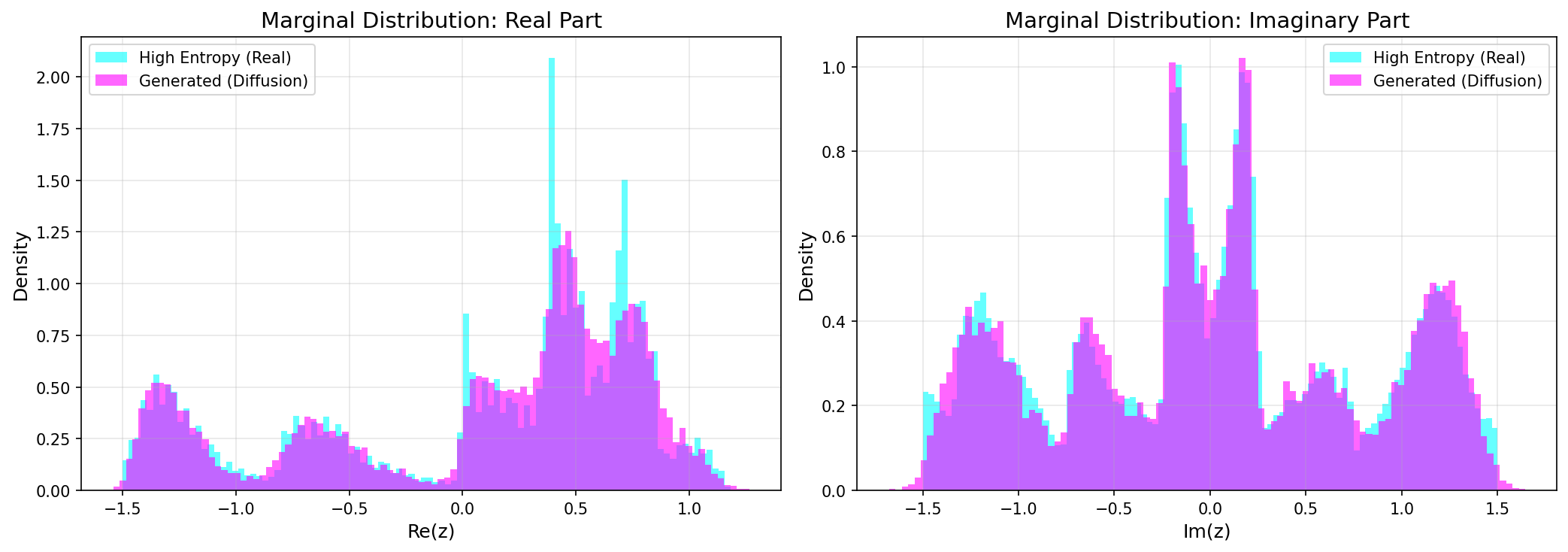}
    \caption{For the Newton fractal, marginal distributions in histogram form of (a) the high entropy samples corresponding to the decision boundary of the classifier plotted along with (b) the marginal distributions of the SGM generated samples.}
    \label{fig:fractal-marginals}
\end{figure}

Once the framework was implemented and executed, the marginal densities of the high entropy points corresponding to the decision boundary of the classifier were compared with the marginal densities of the SGM generated samples. Two metrics were used to quantitatively compare the two sets: Wasserstein distance, used generally to measure the "work" required to transform one density into another \cite{givens1984class}, and the Chamfer distance, the average of nearest‑neighbor distances from each point in one set to the closest point in the other set \cite{fan2017point}. These quantities effectively measure the difference in density and difference in coverage of the two sets, respectively, and the averages of the distances between the two marginals (real and complex) are reported in Table \ref{tab:fractal_metrics}. Figure \ref{fig:fractal-marginals} displays the marginal distributions plotted against each other and qualitatively compared.

\begin{table}
	\caption{Marginal Distribution Comparison Metrics for the Newton Fractal}
	\centering
	\begin{tabular}{lll}
		\toprule
		%\multicolumn{2}{c}{Part}                   \\
		\cmidrule(r){1-2}
		  Metric    & Value \\
		\midrule
		  Chamfer Distance                & 0.0115    \\
		  Sliced Wasserstein Distance     & 0.0493    \\
		\bottomrule
	\end{tabular}
	\label{tab:fractal_metrics}
\end{table}

\subsection{Sensitivity Analysis of the Entropy Threshold with a Continuous Stirred Tank Reactor (CSTR) Example}

A critical consideration in the proposed framework is the selection of the entropy percentile threshold used to identify points near the separatrix. The 90th percentile threshold is chosen in the analysis throughout this work as a reasonable illustration, but depending on the desired quantity and accuracy of the output, other threshold values may be warranted. We performed a sensitivity analysis to examine how variations in the entropy threshold affect the extracted high-entropy point set and to identify regions of stability in the threshold parameter space.

Consider the non-isothermal continuous stirred tank reactor (CSTR) model studied by Uppal, Ray, and Poore \cite{UPPAL1974967}, given by the functional dimensionless equations
\begin{align}
\frac{dx_1}{dt}&=-x_1+Da(1-x_1)\exp({x_2})=f_1(x_1,x_2)
\label{eq:cstr_1}\\
\frac{dx_2}{dt}&=-x_2+BDa(1-x_1)\exp({x_2})-\beta(x_2)=f_2(x_1,x_2) \label{eq:cstr_2}
\end{align}
which exhibit multiple steady states arising from the interplay between reaction kinetics dependent on temperature, concentration, and other factors. Characterizing this separatrix is important for reactor design and control, as it can delineate safe operating regions and determine startup trajectories that avoid extinction or runaway reactions. We investigate this system at the parameter values $B, Da, \beta = 16.2, 0.12823, 3.0$, which exhibit a stable steady state surrounded by an unstable limit cycle (the separatrix), surrounded by a larger stable limit cycle. Following the framework, 40,000 uniformly sampled initial conditions in the range of $x_1,x_2 = [0,1],[0,10].$ The initial conditions were then integrated using equations \ref{eq:cstr_1} and \ref{eq:cstr_2}, and because the stable steady state lies close to the unstable limit cycle, the basins of attraction were labeled by {\em time to convergence} rather than location in phase space. Small perturbations near the steady state can lead to qualitatively different transient behaviors, despite originating from nearly identical regions in phase space, making this separatrix vanishingly small and difficult for the network to identify when using the traditional labels. This grid of initial conditions labeled by basin based on time to convergence of their trajectories were used to train the neural network classifier. Once trained, the classifier evaluated a new finer mesh grid of 160,000 (400x400) samples. 

The sensitivity analysis evaluates thresholds ranging from the 75th to 99th percentile of the entropy distribution. For each candidate threshold, we compute several diagnostic quantities: the number of points exceeding the threshold, the spatial coverage of the resulting point set as a fraction of the domain, the centroid location of the extracted points, and the spatial spread characterized by the standard deviation in each coordinate direction. From these quantities, we derive a combined instability score that measures the sensitivity of the results to small perturbations in the threshold. Specifically, the instability score incorporates the normalized rate of change in point count with respect to the percentile threshold and the magnitude of centroid displacement as the threshold varies. Regions where this score is minimized correspond to threshold values where the extracted separatrix approximation is most stable with respect to the threshold hyperparameter. Mathematically, the score is defined as follows: 

Let $N(p)$ denote the number of grid points whose entropy exceeds the $p$-th percentile threshold, and let $\mu_{x_1}(p)$, $\mu_{x_2}(p)$ denote the coordinates of the centroid of that high-entropy point set, each regarded as continuous functions of the percentile $p$. The combined instability score $S(p)$, which quantifies how sensitive the detected separatrix is to the choice of threshold, is defined as:

\begin{equation}
S(p) \;=\; \frac{\left|\dfrac{dN(p)}{dp}\right|}{\displaystyle\max_{p}\left|\dfrac{dN(p)}{dp}\right|}
\;+\;
\frac{\sqrt{\left(\dfrac{d\mu_{x_1}(p)}{dp}\right)^{2} + \left(\dfrac{d\mu_{x_2}(p)}{dp}\right)^{2}}}
{\displaystyle\max_{p}\sqrt{\left(\dfrac{d\mu_{x_1}(p)}{dp}\right)^{2} + \left(\dfrac{d\mu_{x_2}(p)}{dp}\right)^{2}}}
\label{eq:inst_score}
\end{equation}

where the first term captures the normalized rate of change of the number of detected high-entropy points with respect to the threshold percentile (sensitivity of the point count), and the second term captures the normalized magnitude of the rate of change of the centroid position (sensitivity of the spatial location of the detected region). Both derivatives are normalized by their respective maxima over the sampled percentile range so that each term lies in $[0,1]$, and the two are summed to produce a single scalar measure of overall instability at each threshold $p$. Lower values of $S(p)$ indicate percentile choices for which the separatrix detection is more robust to small perturbations in the threshold.

\begin{figure}[H]
    \centering
    \includegraphics[width=0.9\linewidth]{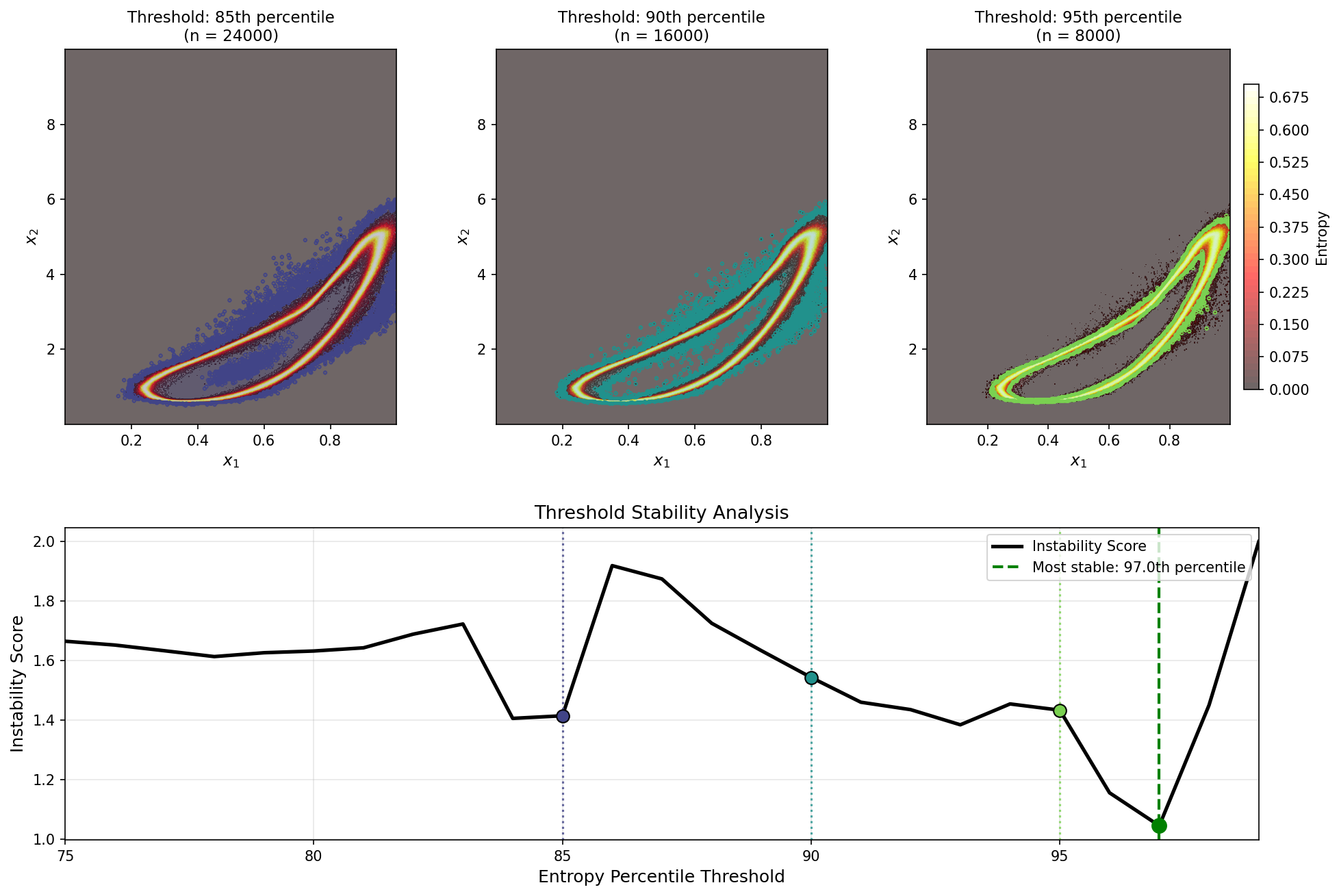}
    \caption{Sensitivity analysis of the entropy threshold selection for the CSTR system. \textbf{Top row:} High-entropy regions identified at the 85th, 90th, and 95th percentile thresholds of the instability score defined by equation \ref{eq:inst_score} colored as purple, cyan, and light green, respectively, plotted with the entropy field computed by the MC Dropout classifier shaded from dark red to white. \textbf{Bottom:} Instability score as a function of the percentile threshold; lower values indicate greater stability with respect to threshold perturbations.
    The green dashed line indicates the most stable threshold, while colored dotted lines correspond to the thresholds visualized in the top row.}
    \label{fig:cstr_sensitivity}
\end{figure}

Figure I\ref{fig:cstr_sensitivity} presents the results of this analysis. The top row displays the high-entropy regions identified at three representative threshold values (85th, 90th, and 95th percentiles), overlaid on the entropy field computed by the MC Dropout classifier. As the threshold increases, the extracted region contracts to a tighter estimation of the separatrix where classification uncertainty is highest. The bottom panel shows the instability score as a function of the percentile threshold. The vertical dashed lines indicate the thresholds corresponding to the top row visualizations, while the green dashed line marks the most stable threshold identified by the analysis. The 97th percentile of points gives the optimal instability score, indicating that it would be the appropriate threshold for the most exact point-wise approximation of the separatrix in this example, trading off less SGM training data for estimation accuracy.  

\begin{figure}[H]
    \centering
    \includegraphics[width=0.4\linewidth]{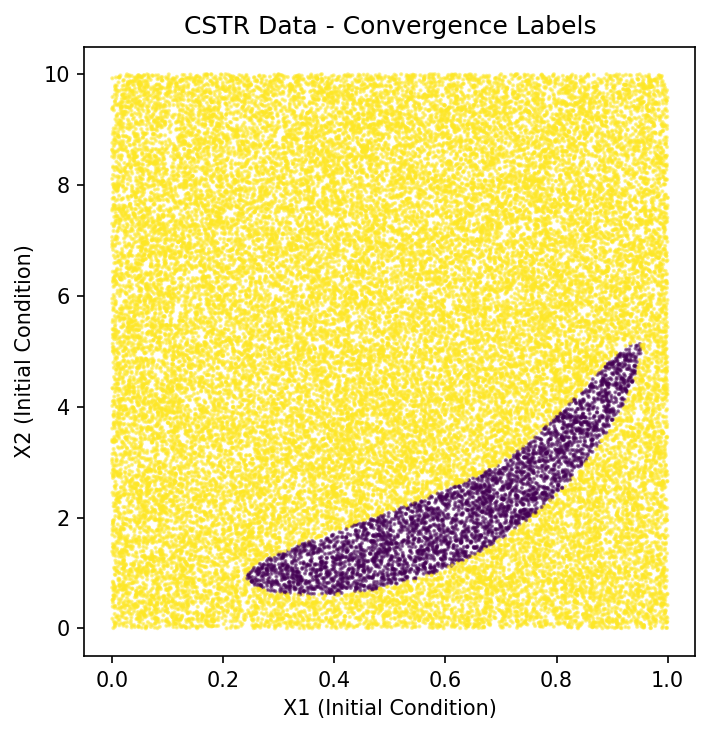}
    \caption{Uniformly sampled initial conditions of the CSTR model equations defined by equations \ref{eq:cstr_1} and \ref{eq:cstr_2} colored by binary label of the corresponding attractor identified by time to convergence (purple for the stable steady state and yellow for the stable limit cycle).}
    \label{fig:cstr_training}
\end{figure}

\begin{figure}[H]
    \centering
    \includegraphics[width=0.9\linewidth]{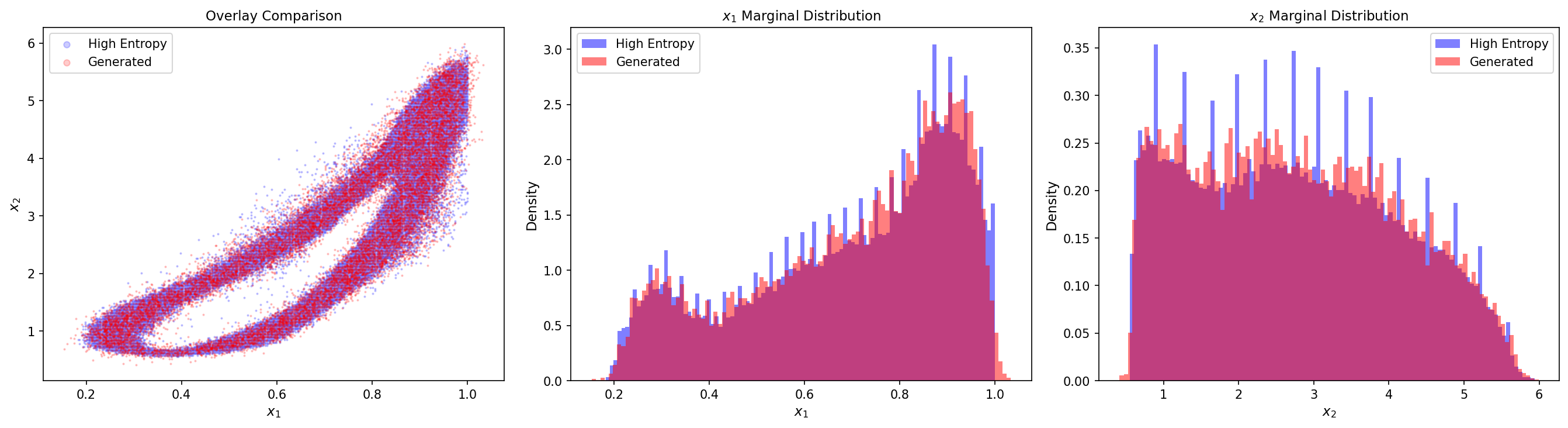}
    \caption{Comparison of high-entropy points and the SGM-generated samples for the CSTR system. \textbf{Left:} Scatter plot overlay of the high-entropy training points (blue) extracted from the MC Dropout classifier at the 90th percentile threshold and the samples (red) generated by the trained score-based generative model. The spatial agreement between the two distributions indicates that the SGM has successfully learned to sample from the separatrix. \textbf{Middle:} Marginal density distributions along the $x_1$ coordinate (dimensionless concentration), showing close agreement between the high-entropy and the generated samples. \textbf{Right:} Marginal density distributions along the $x_2$ coordinate. The strong overlap in both marginal distributions demonstrates that the generative model accurately captures the statistical structure of the classifier's decision boundary corresponding to the dynamical separatrix. The ``peaks" in the high entropy marginals are a product of their uniform sampling.}
    \label{fig:cstr_scatter_marginals}
\end{figure}

Maintaining consistency with the rest of this work, rather than optimizing each example's cutoff distance, we demonstrate the method with a uniform entropy cutoff. The top ten percent of the evaluated samples with respect to their entropy predicted by the classifier (16,000 samples) were retained for SGM training. Once trained on the high entropy points, the SGM then generated 16,000 new samples for comparison with the high entropy training dataset that effectively approximates the decision boundary (and thus the separatrix) of the system. Figure \ref{fig:cstr_training} displays the training data given by the classifier, and Figure \ref{fig:cstr_scatter_marginals} displays the scatter plot of the high entropy points and SGM generated samples as well as their marginal distributions. The Wasserstein and Chamfer distance metrics were again computed to quantify the difference in density and difference in coverage of the two sets and the averages of the distances between the two marginals (real and complex). Their values are $0.0572$ and $0.0088$, respectively.

\subsection{A Comparison of Methods with the Lorenz System}

The Lorenz system, originally derived as a simplified model of atmospheric convection, is a three-dimensional autonomous system given by
\begin{align}
    \frac{dx}{dt} &= \sigma(y - x), \\
    \frac{dy}{dt} &= x(\rho - z) - y, \\
    \frac{dz}{dt} &= xy - \beta z,
\end{align}
where $(x, y, z) \in \mathbb{R}^3$ represent the system state and $\sigma$, $\rho$, and $\beta$ are parameters. We consider the parameter values $\sigma = 10$, $\rho = 20$, and $\beta = 8/3$. At these parameters, which lie below the critical value $\rho_c \approx 24.74$ for the onset of chaos, the system exhibits bistability between two stable fixed points $C^{\pm} = (\pm\sqrt{\beta(\rho-1)}, \pm\sqrt{\beta(\rho-1)}, \rho-1)$, while the origin remains an unstable saddle with a two-dimensional stable manifold and a one-dimensional unstable manifold. The stable manifold $W^s(\mathbf{0})$ of the origin forms the separatrix, a two-dimensional surface that divides the phase space into two basins of attraction corresponding to $C^+$ and $C^-$. This configuration provides an interesting test case for three-dimensional separatrix reconstruction in a system with well-defined multistability, a geometrically complex basin boundary structure, and chaotic transient behavior. This separatrix is the 2D stable manifold of the origin of the 3D Lorenz system described by the above parameter values, which we will briefly call the "2D Lorenz Manifold" from now on.

\begin{figure}[H]
    \centering
    \includegraphics[width=0.9\linewidth]{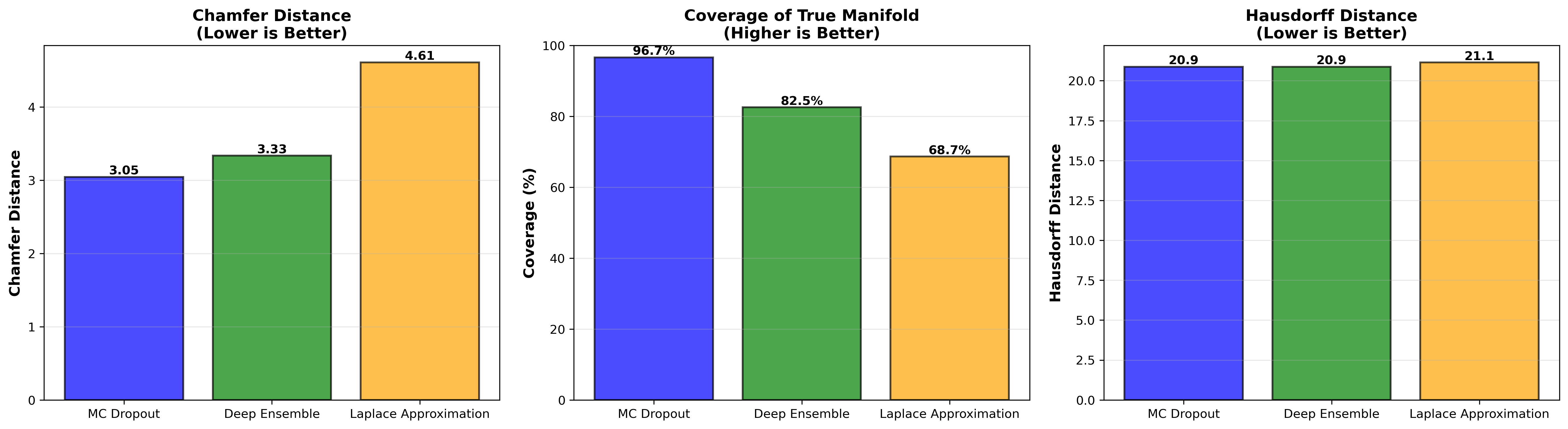}
    \caption{For the Lorenz System, Comparison of high entropy points from three different methods (MC dropout, deep ensemble averaging, and Laplace approximation) versus a sampled manifold obtained via SCIGMA. The three metrics used are Chamfer distance, coverage, and Hausdorff distance.}
    \label{fig:lorenz_metrics}
\end{figure}

To assess the effectiveness of different uncertainty quantification approaches for separatrix identification, we compared three methods for estimating predictive uncertainty from the trained classifier: Monte Carlo (MC) dropout, deep ensemble averaging \cite{lakshminarayanan2017simple}, and Laplace approximation \cite{mackay1992practical}. For each method, we identified high-uncertainty regions and evaluated their geometric correspondence to the ground truth separatrix computed using the SCIGMA~\cite{scigma} continuation software.\footnote{The ground truth separatrix was computed using SCIGMA, a software package for invariant-manifold computation and visualization~\cite{scigma}. We used the December 2025 macOS ARM release with the \texttt{mstable} continuation commands, initialized from the unstable saddle at $(0,0,0)$. Numerical integration used $dt=0.005$; AUTO continuation used $ds=0.01$, $ds_{\min}=10^{-6}$, and $ds_{\max}=1$, with all other parameters set to their defaults.}
Performance was quantified using three metrics: coverage (the fraction of true separatrix points within a specified distance of high-uncertainty regions), Chamfer distance (measuring average nearest-neighbor distances between predicted and true boundary points in both directions, described and cited in previous sections), and Hausdorff distance (capturing worst-case deviations) \cite{huttenlocher2002comparing}. MC dropout outperformed the alternatives in coverage and Chamfer distance while staying comparatively equal or better with the other methods in Hausdorff distance, as detailed in Figure \ref{fig:lorenz_metrics}. The superior performance of MC dropout likely stems from its ability to capture local uncertainty variations more effectively than the global uncertainty estimates provided by methods such as Laplace approximation, while requiring substantially less computational cost than training full ensembles.

\begin{figure}
    \centering
    \includegraphics[width=0.75\linewidth]{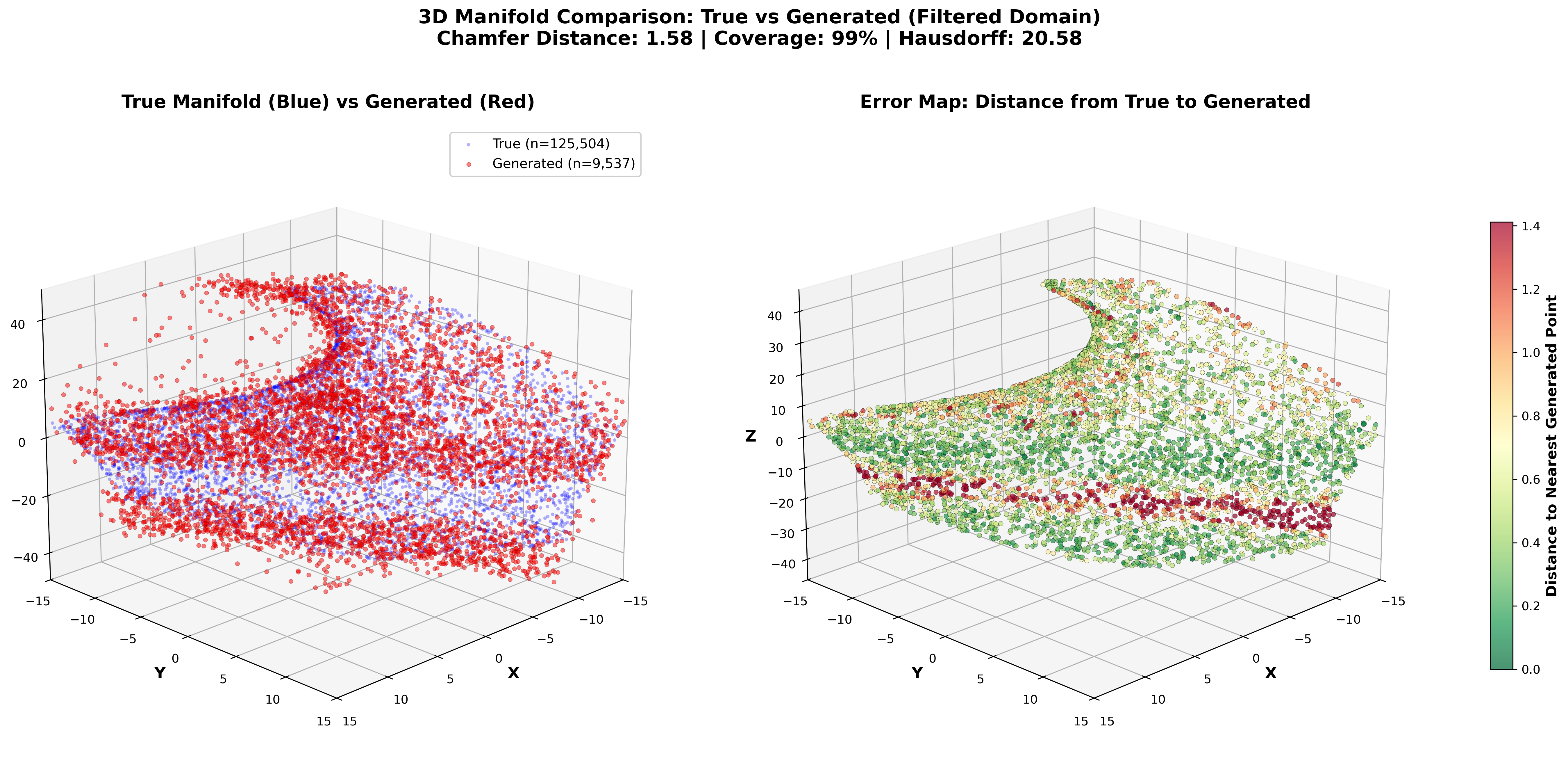}
    \caption{Three-dimensional plots of the approximated 2D Lorenz Manifold. Left: An overlaid comparison of the generated approximation of the manifold using the generative method plotted as red points and the sampled manifold found via SCIGMA in blue points. Right: the SCIGMA calculated manifold points, colored by distance to the approximated manifold (effectively the Chamfer distance).}
    \label{fig:lorenz_3d}
\end{figure}

\begin{figure}[H]
    \centering
    \includegraphics[width=0.75\linewidth]{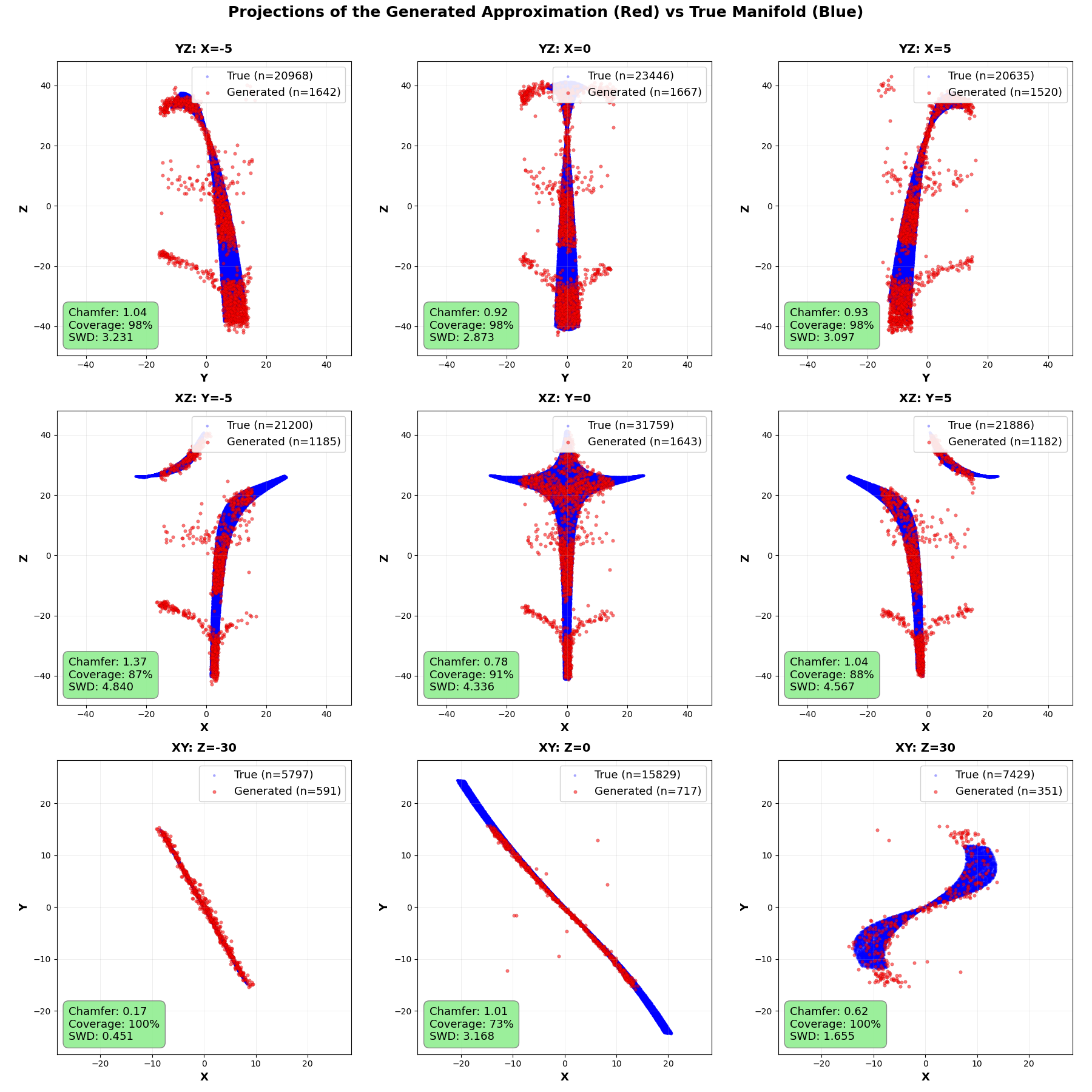}
    \caption{Sets of two dimensional projections from various sections of the separatrix calculated using traditional methods (plotted in blue), versus the generated points that approximate the separatrix (maked in red). Inset metrics describing the Chamfer distance, coverage, and sliced Wasserstein distance of each projection are included in each plot.}
    \label{fig:lorenz_projections}
\end{figure}

Using MC dropout as the uncertainty quantification method of choice, we applied the complete framework (classification, boundary identification via high-uncertainty regions, and generative modeling on boundary data) to reconstruct the separatrix of the Lorenz system. The final output from the score-based generative model, comprising samples along the reconstructed boundary, was compared against the ground truth manifold. The generative model substantially improved the quality of the reconstruction relative to the classifier alone: the Chamfer distance decreased from 3.05 to 1.58 (a 48.2\% reduction) and coverage increased from 96.7\% to 99\%. The Hausdorff distance improved slightly (from 20.9 to 20.58), consistent with the low variation seen between the UQ methods. Qualitatively, the generated samples captured the fine-scale geometric features of the separatrix. While this is demonstrated in the improved metrics and not readily apparent from the three dimensional point clouds in Figure \ref{fig:lorenz_3d}, this can be seen when looking at two dimensional projections of the ground truth manifold versus the approximate manifold from the generative model, demonstrated in Figure \ref{fig:lorenz_projections}. These projections do not exactly approximate the true calculated separatrix due to the large sampled region and inherent complexity of the manifold, but with iterations on specific regions, they would approach an exact approximation. These results demonstrate that the integrated classification-generation approach produces separatrix reconstructions with fidelity comparable to specialized continuation methods, while offering advantages in computational efficiency and applicability to systems where continuation methods face challenges, such as when the equations to a system are not known \emph{a priori}.

\section{Conclusion}

This work has introduced a data-driven framework for computationally approximating separatrices in multistable dynamical systems by combining supervised classification with generative modeling. The methodology addresses fundamental sampling challenges inherent in characterizing basin boundaries: these structures govern transitions between stable states yet remain poorly sampled in standard numerical simulations due to their repelling behavior. We have demonstrated the framework's effectiveness on three representative dynamical systems. Across these examples, the method successfully identified and reconstructed basin boundaries with fidelity comparable to traditional techniques while offering advantages in computational efficiency, applicability to high-dimensional systems, and circumvention of required system knowledge. The framework provides particular value for systems where traditional continuation and bisection methods face limitations. By learning directly from trajectory data rather than requiring known equations, it accommodates systems known only through simulation or experiment. The approach naturally extends to high-dimensional phase spaces where visualization fails and manifold continuation becomes prohibitively expensive. Moreover, the generative component addresses a fundamental need: producing representative samples from regions of phase space that standard simulations systematically undersample. This framework demonstrates that modern machine learning, when appropriately integrated with dynamical systems theory, provides powerful capabilities for characterizing global geometric structures in complex nonlinear systems by enabling a data-driven understanding of critical transitions that govern system behavior.

\subsection{Future Work}

The future work will focus on augmenting this method with an iterative and active learning framework to mitigate its existing limitations. \textit{Isolated high-uncertainty regions} can emerge in sparsely sampled areas far from any true separatrix, leading to false positive identification of spurious boundaries. These arise when the network, lacking local training data, defaults to uncertain predictions while still being far from any dynamically significant structure. Several promising directions exist for extending this methodology and mitigating this limitation. First, systematic evaluation of the framework's performance in sparse data regimes would address a critical practical constraint. Understanding how reconstruction quality degrades with reduced training data and identifying minimum data requirements for reliable separatrix identification would guide deployment in data-limited contexts. Developing uncertainty quantification measures that indicate when additional data is needed would further enhance practical utility, and some readily available metrics already exist, e.g., basin entropy~\cite{Daza2016}. Additionally, active learning strategies offer substantial potential for improving sample efficiency. Current implementation treats initial condition sampling as uniform or predetermined, but an adaptive approach could iteratively select the most informative regions for new trajectory generation. 
In future work we plan to utilize functionality offered by the software DynamicalSystems.jl, which provides generic functions for labeling trajectories according to their basin membership~\cite{Datseris2022BasinsAttraction}. This will allow for an iterative refinement of separatrices, iteratively probing high uncertainty regions with denser sampling. 
Finally, integration with optimal experimental design principles could formalize selection criteria to maximize information gain about basin boundary geometry per simulation. Further tests will be performed for the improved methods, including multi-seed robustness tests and improvements on our existing threshold sensitivity studies.

\bibliographystyle{unsrtnat}
\bibliography{references}

@misc{songdiffmodels,
  doi = {10.48550/ARXIV.2011.13456},
  
  url = {https://arxiv.org/abs/2011.13456},
  
  author = {Song, Yang and Sohl-Dickstein, Jascha and Kingma, Diederik P. and Kumar, Abhishek and Ermon, Stefano and Poole, Ben},
  
  title = {Score-Based Generative Modeling through Stochastic Differential Equations},
  
  publisher = {arXiv},
  
  year = {2020},
  
  copyright = {arXiv.org perpetual, non-exclusive license}
}

@misc{diffmodels2,
  doi = {10.48550/ARXIV.2006.09011},
  
  url = {https://arxiv.org/abs/2006.09011},
  
  author = {Song, Yang and Ermon, Stefano},
  
  title = {Improved Techniques for Training Score-Based Generative Models},
  
  publisher = {arXiv},
  
  year = {2020},
  
  copyright = {arXiv.org perpetual, non-exclusive license}
}

@article{GANs_closures,
  title={Gans and closures: Micro-macro consistency in multiscale modeling},
  author={Crabtree, Ellis R and Bello-Rivas, Juan M and Ferguson, Andrew L and Kevrekidis, Ioannis G},
  journal={Multiscale modeling \& simulation},
  volume={21},
  number={3},
  pages={1122--1146},
  year={2023},
  publisher={SIAM}
}

@inproceedings{gal2016dropout,
  title={Dropout as a bayesian approximation: Representing model uncertainty in deep learning},
  author={Gal, Yarin and Ghahramani, Zoubin},
  booktitle={international conference on machine learning},
  pages={1050--1059},
  year={2016},
  organization={PMLR}
}

@article{sleeman2023generative,
  title={A generative adversarial network for climate tipping point discovery (tip-gan)},
  author={Sleeman, Jennifer and Chung, David and Gnanadesikan, Anand and Brett, Jay and Kevrekidis, Yannis and Hughes, Marisa and Haine, Thomas and Pradal, Marie-Aude and Gelderloos, Renske and Ashcraft, Chace and others},
  journal={arXiv preprint arXiv:2302.10274},
  year={2023}
}

@misc{VAE,
  doi = {10.48550/ARXIV.1312.6114},
  
  url = {https://arxiv.org/abs/1312.6114},
  
  author = {Kingma, Diederik P and Welling, Max},
  
  title = {Auto-Encoding Variational Bayes},
  
  publisher = {arXiv},
  
  year = {2013},
  
  copyright = {arXiv.org perpetual, non-exclusive license}
}

@article{goodfellow2014,
	title = {Generative {Adversarial} {Networks}},
	url = {http://arxiv.org/abs/1406.2661},
	urldate = {2021-03-30},
	journal = {arXiv:1406.2661 [cs, stat]},
	author = {Goodfellow, I. J. and Pouget-Abadie, J. and Mirza, M. and Xu, B. and Warde-Farley, D. and Ozair, S. and Courville, A. and Bengio, Y.},
	month = jun,
	year = {2014},
	note = {arXiv: 1406.2661}
}

@article{scheffer2001catastrophic,
  title={Catastrophic shifts in ecosystems},
  author={Scheffer, Marten and Carpenter, Steve and Foley, Jonathan A and Folke, Carl and Walker, Brian},
  journal={Nature},
  volume={413},
  number={6856},
  pages={591--596},
  year={2001},
  publisher={Nature Publishing Group UK London}
}

@article{ELNASHAIE19931,
title = {Bifurcation, instability and chaos in fluidized bed catalytic reactors with consecutive exothermic chemical reactions},
journal = {Chaos, Solitons and Fractals},
volume = {3},
number = {1},
pages = {1-33},
year = {1993},
issn = {0960-0779},
doi = {https://doi.org/10.1016/0960-0779(93)90037-2},
url = {https://www.sciencedirect.com/science/article/pii/0960077993900372},
author = {S.S. Elnashaie and M.E. Abashar and F.A. Teymour}
}

@article{UPPAL1974967,
title = {On the dynamic behavior of continuous stirred tank reactors},
journal = {Chemical Engineering Science},
volume = {29},
number = {4},
pages = {967-985},
year = {1974},
issn = {0009-2509},
doi = {https://doi.org/10.1016/0009-2509(74)80089-8},
url = {https://www.sciencedirect.com/science/article/pii/0009250974800898},
author = {A. Uppal and W.H. Ray and A.B. Poore}
}

@article{krauskopf2005survey,
  title={A survey of methods for computing (un) stable manifolds of vector fields},
  author={Krauskopf, Bernd and Osinga, Hinke M and Doedel, Eusebius J and Henderson, Michael E and Guckenheimer, John and Vladimirsky, Alexander and Dellnitz, Michael and Junge, Oliver},
  journal={International Journal of Bifurcation and Chaos},
  volume={15},
  number={03},
  pages={763--791},
  year={2005},
  publisher={World Scientific}
}

@article{skufca2006edge,
  title={Edge of chaos in a parallel shear flow},
  author={Skufca, Joseph D and Yorke, James A and Eckhardt, Bruno},
  journal={Physical review letters},
  volume={96},
  number={17},
  pages={174101},
  year={2006},
  publisher={APS}
}

@article{feudel1997multistability,
  title={Multistability and the control of complexity},
  author={Feudel, Ulrike and Grebogi, Celso},
  journal={Chaos: An Interdisciplinary Journal of Nonlinear Science},
  volume={7},
  number={4},
  pages={597--604},
  year={1997},
  publisher={American Institute of Physics}
}

@article{crabtree2024micro,
  title={Micro-macro consistency in multiscale modeling: Score-based model assisted sampling of fast/slow dynamical systems},
  author={Crabtree, ER and Bello-Rivas, JM and Kevrekidis, IG},
  journal={Chaos: An Interdisciplinary Journal of Nonlinear Science},
  volume={34},
  number={5},
  year={2024},
  publisher={AIP Publishing}
}

@article{giovanis2025generative,
  title={Generative learning of densities on manifolds},
  author={Giovanis, Dimitris G and Crabtree, Ellis and Ghanem, Roger G and Kevrekidis, Ioannis G},
  journal={Computer Methods in Applied Mechanics and Engineering},
  volume={446},
  pages={118266},
  year={2025},
  publisher={Elsevier}
}

@article{crabtree2025generative,
  title={Generative learning for slow manifolds and bifurcation diagrams},
  author={Crabtree, Ellis R and Giovanis, Dimitris G and Evangelou, Nikolaos and Bello-Rivas, Juan M and Kevrekidis, Ioannis G},
  journal={Computers \& Chemical Engineering},
  pages={109544},
  year={2025},
  publisher={Elsevier}
}

@misc{diffmanifolds,
  doi = {10.48550/ARXIV.2206.01018},
  
  url = {https://arxiv.org/abs/2206.01018},
  
  author = {Pidstrigach, Jakiw},
  
  title = {Score-Based Generative Models Detect Manifolds},
  
  publisher = {arXiv},
  
  year = {2022},
  
  copyright = {arXiv.org perpetual, non-exclusive license}
}

@article{givens1984class,
  title={A class of Wasserstein metrics for probability distributions.},
  author={Givens, Clark R and Shortt, Rae Michael},
  journal={Michigan Mathematical Journal},
  volume={31},
  number={2},
  pages={231--240},
  year={1984},
  publisher={University of Michigan, Department of Mathematics}
}

@inproceedings{fan2017point,
  title={A point set generation network for 3d object reconstruction from a single image},
  author={Fan, Haoqiang and Su, Hao and Guibas, Leonidas J},
  booktitle={Proceedings of the IEEE conference on computer vision and pattern recognition},
  pages={605--613},
  year={2017}
}

@article{huttenlocher2002comparing,
  title={Comparing images using the Hausdorff distance},
  author={Huttenlocher, Daniel P and Klanderman, Gregory A. and Rucklidge, William J},
  journal={IEEE Transactions on pattern analysis and machine intelligence},
  volume={15},
  number={9},
  pages={850--863},
  year={2002},
  publisher={IEEE}
}

@article{mackay1992practical,
  title={A practical Bayesian framework for backpropagation networks},
  author={MacKay, David JC},
  journal={Neural computation},
  volume={4},
  number={3},
  pages={448--472},
  year={1992},
  publisher={MIT Press One Rogers Street, Cambridge, MA 02142-1209, USA journals-info~…}
}

@article{lakshminarayanan2017simple,
  title={Simple and scalable predictive uncertainty estimation using deep ensembles},
  author={Lakshminarayanan, Balaji and Pritzel, Alexander and Blundell, Charles},
  journal={Advances in neural information processing systems},
  volume={30},
  year={2017}
}

@misc{scigma,
  title        = {{SCIGMA}: Stability Computations and Interactive Graphics for invariant Manifold Analysis},
  author       = {Kevrekidis, I. G. and Jolly, M. S. and Taylor, M. A. and Johnson, M. E. and H{\"o}lzel, Robert},
  year         = {2025},
  howpublished = {\url{https://scigma.org/}},
}

@book{Pisarchik2022,
   author = {Alexander N Pisarchik and Alexander E Hramov},
   city = {Cham},
   doi = {10.1007/978-3-030-98396-3},
   isbn = {978-3-030-98395-6},
   publisher = {Springer International Publishing},
   title = {Multistability in Physical and Living Systems},
   url = {https://link.springer.com/10.1007/978-3-030-98396-3},
   year = {2022}
}

@article{Feudel2018MultistabilityTippingMathematics,
	title = {Multistability and tipping: {From} mathematics and physics to climate and brain—{Minireview} and preface to the focus issue},
	volume = {28},
	issn = {1054-1500, 1089-7682},
	shorttitle = {Multistability and tipping},
	doi = {10.1063/1.5027718},
	number = {3},
	journal = {Chaos: An Interdisciplinary Journal of Nonlinear Science},
	author = {Feudel, Ulrike and Pisarchik, Alexander N. and Showalter, Kenneth},
	month = mar,
	year = {2018},
	pages = {033501},
}

@book{Datseris2022NonlinearDynamicsJulia,
	address = {Cham},
	series = {Undergraduate {Lecture} {Notes} in {Physics}},
	title = {Nonlinear {Dynamics}: {A} {Concise} {Introduction} {Interlaced} with {Code}},
	copyright = {https://www.springer.com/tdm},
	isbn = {978-3-030-91031-0 978-3-030-91032-7},
	shorttitle = {Nonlinear {Dynamics}},
	publisher = {Springer International Publishing},
	author = {Datseris, George and Parlitz, Ulrich},
	year = {2022},
	doi = {10.1007/978-3-030-91032-7},
}

@article{Datseris2022BasinsAttraction,
	title = {Effortless estimation of basins of attraction},
	volume = {32},
	issn = {1054-1500, 1089-7682},
	doi = {10.1063/5.0076568},
	number = {2},
	journal = {Chaos: An Interdisciplinary Journal of Nonlinear Science},
	author = {Datseris, George and Wagemakers, Alexandre},
	month = feb,
	year = {2022},
	pages = {023104},
}

@article{Datseris2023FrameworkGlobalStability,
	title = {Framework for global stability analysis of dynamical systems},
	volume = {33},
	issn = {1054-1500, 1089-7682},
	doi = {10.1063/5.0159675},
	number = {7},
	urldate = {2024-09-02},
	journal = {Chaos: An Interdisciplinary Journal of Nonlinear Science},
	author = {Datseris, George and Luiz Rossi, Kalel and Wagemakers, Alexandre},
	month = jul,
	year = {2023},
	pages = {073151},
}

@misc{Boerner2025,
  doi = {10.48550/ARXIV.2504.20002},
  url = {https://arxiv.org/abs/2504.20002},
  author = {B\"{o}rner,  Reyk and Mehling,  Oliver and von Hardenberg,  Jost and Lucarini,  Valerio},
  title = {Global stability of the Atlantic overturning circulation: Edge state,  long transients and boundary crisis under CO$_2$ forcing},
  publisher = {arXiv},
  year = {2025},
  copyright = {Creative Commons Attribution 4.0 International}
}

@article{Armiyoon2014,
  title = {A novel method to identify boundaries of basins of attraction in a dynamical system using Lyapunov exponents and Monte Carlo techniques},
  volume = {79},
  ISSN = {1573-269X},
  url = {http://dx.doi.org/10.1007/s11071-014-1663-z},
  DOI = {10.1007/s11071-014-1663-z},
  number = {1},
  journal = {Nonlinear Dynamics},
  publisher = {Springer Science and Business Media LLC},
  author = {Armiyoon,  Ali Reza and Wu,  Christine Q.},
  year = {2014},
  pages = {275–293}
}

@article{johnson1997two,
  title={Two-dimensional invariant manifolds and global bifurcations: some approximation and visualization studies},
  author={Johnson, Mark E and Jolly, Michael S and Kevrekidis, Ioannis G},
  journal={Numerical Algorithms},
  volume={14},
  number={1},
  pages={125--140},
  year={1997},
  publisher={Springer}
}

@article{kevrekidis1987bifurcations,
author = {Kevrekidis, I. G.},
title = {A numerical study of global bifurcations in chemical dynamics},
journal = {AIChE Journal},
volume = {33},
number = {11},
pages = {1850-1864},
doi = {https://doi.org/10.1002/aic.690331112},
url = {https://aiche.onlinelibrary.wiley.com/doi/abs/10.1002/aic.690331112},
eprint = {https://aiche.onlinelibrary.wiley.com/doi/pdf/10.1002/aic.690331112},
year = {1987}
}

@article{aronson1982,
author = {D. G. Aronson and M. A. Chory and G. R. Hall and R. P. McGehee},
title = {{Bifurcations from an invariant circle for two-parameter families of maps of the plane: a computer-assisted study}},
volume = {83},
journal = {Communications in Mathematical Physics},
number = {3},
publisher = {Springer},
pages = {303 -- 354},
year = {1982},
}

@article{Daza2016,
  title = {Basin entropy: a new tool to analyze uncertainty in dynamical systems},
  volume = {6},
  ISSN = {2045-2322},
  url = {http://dx.doi.org/10.1038/srep31416},
  DOI = {10.1038/srep31416},
  number = {1},
  journal = {Scientific Reports},
  publisher = {Springer Science and Business Media LLC},
  author = {Daza,  Alvar and Wagemakers,  Alexandre and Georgeot,  Bertrand and Guéry-Odelin,  David and Sanjuán,  Miguel A. F.},
  year = {2016},
  month = Aug 
}

\end{document}